\documentclass[conference]{IEEEtran}
\IEEEoverridecommandlockouts

\usepackage{cite}
\usepackage{amsmath,amssymb,amsfonts}
\usepackage{algorithmic}
\usepackage{graphicx}
\usepackage{textcomp}
\usepackage{xcolor}
\usepackage{soul}

\def\BibTeX{{\rm B\kern-.05em{\sc i\kern-.025em b}\kern-.08em
    T\kern-.1667em\lower.7ex\hbox{E}\kern-.125emX}}
\begin{document}

\title{Regional Correlation Aided Mobile Traffic Prediction with Spatiotemporal Deep Learning \\
}

\author{
JeongJun Park$^{1}$,
Lusungu J. Mwasinga$^{2}$,
Huigyu Yang$^{3}$, 
Syed M. Raza\textsuperscript{1*}, 
Duc-Tai Le\textsuperscript{4*},
\\
Moonseong Kim$^{5}$,
Min Young Chung$^{1}$, and
Hyunseung Choo\textsuperscript{1234*}
\\\\
$^{1}$Department of Electrical and Computer Engineering, Sungkyunkwan University, Suwon, Korea
\\
$^{2}$Department of Computer Science and Engineering, Sungkyunkwan University, Suwon, Korea
\\
$^{3}$Department of Superintelligence Engineering, Sungkyunkwan University, Suwon, Korea
\\
$^{4}$College of Computing and Informatics, Sungkyunkwan University, Suwon, Korea
\\
$^{5}$Department of IT Convergence Software, Seoul Theological University, Bucheon, Korea
\\
Emails:\{parkjj1234, s.moh.raza, ldtai, choo\}@skku.edu}
\maketitle

\begin{abstract}
Mobile traffic data in urban regions shows differentiated patterns during different hours of the day. The exploitation of these patterns enables highly accurate mobile traffic prediction for proactive network management. However, recent Deep Learning (DL) driven studies have only exploited spatiotemporal features and have ignored the geographical correlations, causing high complexity and erroneous mobile traffic predictions. This paper addresses these limitations by proposing an enhanced mobile traffic prediction scheme that combines the clustering strategy of daily mobile traffic peak time and novel multi Temporal Convolutional Network with a Long Short Term Memory (multi TCN-LSTM) model. The mobile network cells that exhibit peak traffic during the same hour of the day are clustered together. Our experiments on large-scale real-world mobile traffic data show up to 28\% performance improvement compared to state-of-the-art studies, which confirms the efficacy and viability of the proposed approach.
\end{abstract}


\section{Introduction}

Mobile traffic patterns in urban regions can be primarily classified into commercial and residential categories \cite{he2020graph}. In commercial areas, mobile traffic peaks during work hours, whereas it increases during leisure hours in residential areas. Understanding these patterns enhances precise mobile traffic predictions. Earlier mobile traffic prediction solutions have used time-series Deep Learning (DL) techniques like Recurrent Neural Networks (RNN), Long Short-Term Memory (LSTM) \cite{rau2021forescating}, \cite{wan2022network}. These models are suited for time-series data as they can effectively identify temporal patterns within the data. However, their performance is limited as they do not exploit the spatial characteristics of mobile traffic. To overcome this, recent studies have integrated deep convolutional techniques in time-series models to effectively learn from spatial features \cite{tiwari2022internet,sone2020wireless}. These combined techniques can capture the spatiotemporal characteristics of mobile traffic data and yield more accurate predictions.

The spatial features captured by the convolutional techniques are agnostic to regional characteristics, especially across vast geographical areas. In mobile traffic prediction, clustering is a better technique to harness the geographical attributes where neighboring cells can be clustered together using the k-means algorithm \cite{zhang2021dual}. The assumption that geographical proximity translates into similar traffic patterns is weak and can be handled by clustering based on the similarity coefficient between traffic patterns of different cells \cite{chen2023spatial}. The calculation cost of the Dynamic Time Warping (DTW) algorithm is high because it cannot correlate multiple cells simultaneously. This calls for a lightweight and sophisticated clustering strategy for diverse cells.

This work addresses the limitations of earlier works by proposing a clustering strategy based on geospatial and temporal patterns. In particular, for each cell peak traffic occurrence time during the day is observed, and using Pearson Correlation Coefficient (PCC) similarity among cells is established. Groups of cells with high similarity are clustered together and predicted through a novel multi Temporal Convolutional Network with an LSTM (multi TCN-LSTM) model. The proposed multi TCN-LSTM model improves performance and ensures stability by adding residual convolutional connections (RCCs) to the existing TCN-LSTM structure. Through experimental results, the proposed approach has demonstrated superior performance in analyzing and predicting mobile traffic patterns. In summary, the major contributions of this paper are as follows:

\begin{itemize}
    \item A novel clustering method classifies regions by peak traffic times. The PCC compares trends between regions. This clustering builds a prediction model for each cluster, enhancing mobile traffic pattern understanding.

    \item A proposed multi TCN-LSTM model comprehensively captures the underlying geospatial and temporal patterns in mobile traffic data by combining residual convolutional connections with the existing TCN-LSTM structure.

    \item A thorough evaluation of the proposed approach using a large-scale mobile traffic dataset \cite{barlacchi2015multi}. Experimental results show $\sim$28\% improvement compared to state-of-the-art, showcasing its utility in diverse real-world scenarios.

\end{itemize}

\begin{figure*}[!h]
\includegraphics[width=\textwidth]{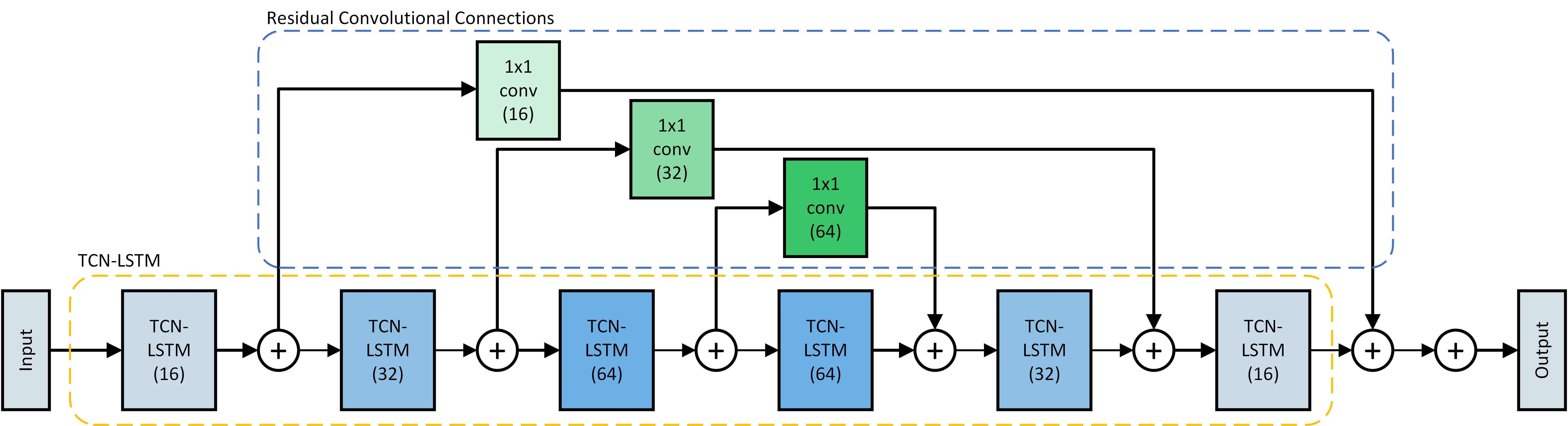}
\caption{multi TCN-LSTM architecture, (16, 32, 64) are number of convolution layer channels.}
\label{fig_1}
\end{figure*}

\section{Enhanced Mobile Traffic Prediction}

\subsection{Multi TCN-LSTM}\label{AA}

The TCN effectively processes time-series data using a 1-dimensional CNN with multiple layers, capturing intricate patterns via casual convolution. This technique focuses on current and past data while excluding future influence. Additionally, TCN dilated convolution broadens the learning scope with optimized computation, and its residual block structure boosts learning speed while addressing the vanishing gradient issue. TCN-LSTM merges the strengths of TCN spatial pattern recognition using a 1-dimensional CNN and LSTM proficiency in temporal learning. This combination effectively captures geospatial and temporal traits in time-series data, crucial for mobile traffic prediction given its fluctuating and intricate patterns. However, the conventional TCN-LSTM may face challenges with long-term dependencies and risk of overfitting due to the model complexity.

The proposed multi TCN-LSTM architecture in Fig.~\ref{fig_1} overcomes the shortcomings of conventional TCN-LSTM. This structure enhances the TCN-LSTM framework by proposing RCCs for improved management of complex mobile traffic data. These  RCCs offer interconnections among the TCN-LSTM layers within the same channel, playing a crucial role in stabilizing and deepening the learning process of the neural network. By reinforcing the flow of features between layers, they effectively resolve the vanishing gradient problem, a common issue in recurrent neural networks. The TCN-LSTM layers consist of two components: a TCN layer specialized in extracting the spatial characteristics of time-series data and an LSTM layer for capturing temporal characteristics. The TCN layer learns the spatial pattern of the given input data, and the learned pattern is subsequently passed to the LSTM layer for temporal prediction. The proposed multi TCN-LSTM architecture in this study demonstrates superior performance in learning and predicting complex spatiotemporal patterns by improving the existing TCN-LSTM framework. It extends beyond merely learning and predicting complex temporal patterns to incorporate multi-dimensional and dynamic spatial patterns as well, thereby providing enhanced analytical power for more complex time-series data.

\begin{figure}[!t]
\begin{minipage}{0.5\textwidth}
\centering
\includegraphics[width=0.95\linewidth]{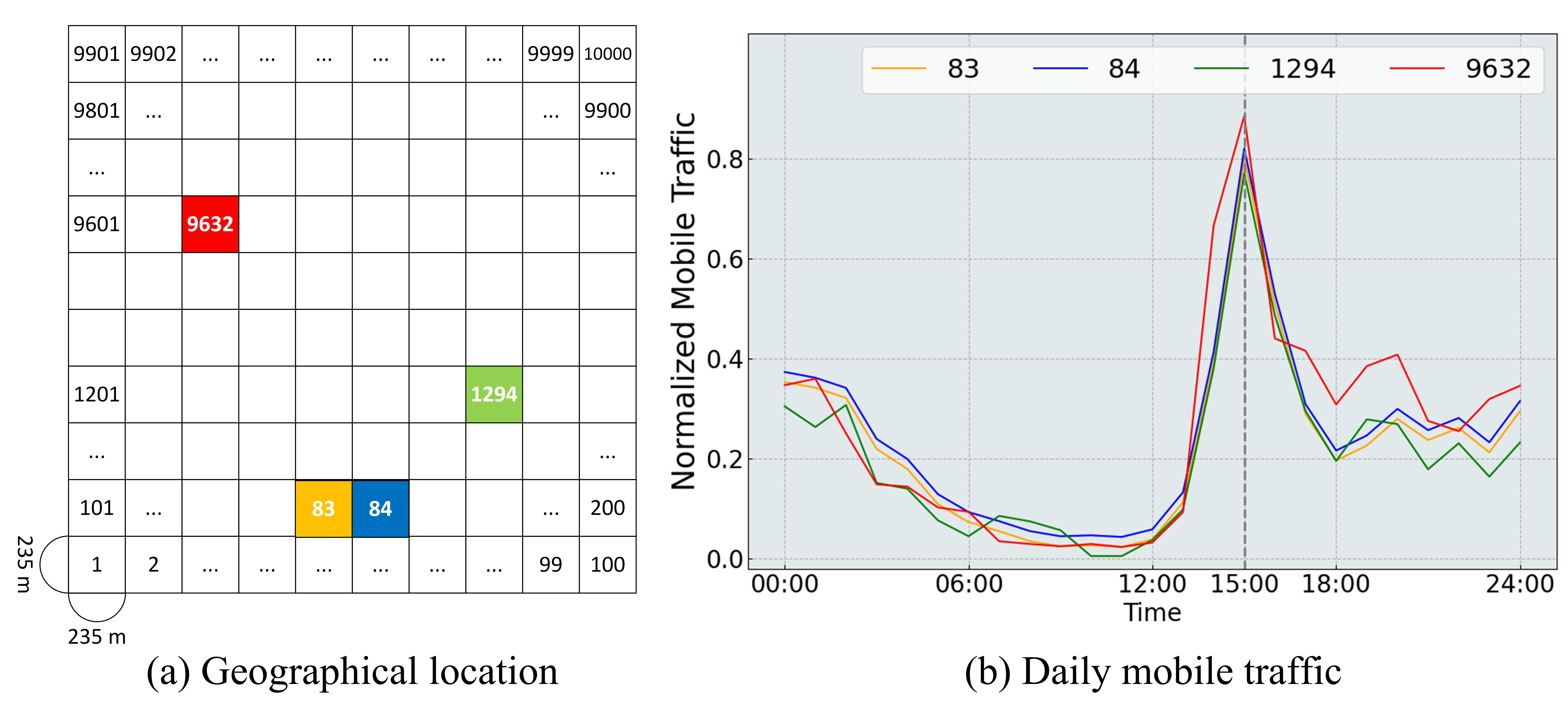}
\caption{Mobile traffic trends based on spatial location and temporal pattern.}
\label{fig_2}
\end{minipage}
\end{figure}

\subsection{Clustering using peak traffic time}\label{AA}

Mobile traffic patterns in a region have a strong geospatial correlation with nearby locations typically showing similar traffic patterns. Fig.~\ref{fig_2} shows neighboring 83 and 84 cells with similar traffic patterns due to their similar regional features. Interestingly, cells 1294 and 9612, though distant, exhibit patterns similar to cells 83 and 84. This signifies the limitation of only utilizing spatial correlation to group regions for traffic prediction. To this end, we analyze the mobile traffic data of cells using their peak traffic occurrence times. As shown in Fig.~\ref{fig_2}, irrespective of spatial location, cells 83, 84, 1294, and 9632 all show peak traffic occurring at 15:00. Hence, only by observing peak traffic occurrence time, geospatial correlations between cells can be established. 

Given these characteristics, this study proposes a clustering method that consists of two main steps. The first step accumulates the traffic data in each hour over a 24-hour period and determines the hour that experiences the peak traffic for each cell. This step for each cell is then repeatedly performed over 20 days, and the most frequently occurring peak traffic hour for a cell is selected as its representative peak traffic time. Cells with identical peak traffic times are then grouped together, resulting in 24 groups. Notably, another advantage of using peak traffic time as a clustering criterion is that it reflects unique lifestyle patterns and attributes of a region. 

The resulting 24 groups require 24 individual DL models to predict network traffic. However, this approach is computationally expensive and requires a mechanism to further combine similar groups. To this end, PCC is calculated between the groups in the second step, which is a widely used statistical tool to measure the strength and direction of linear relationships among two variables \cite{li2020prediction}. Its value ranges between -1 and 1, with 1 indicating a strong positive correlation, and -1 indicating a strong negative correlation. The PCC $r$ among groups $x$ and $y$ is calculated as: 

\begin{equation}
    r_{x,y} = \frac{\sum_{j=0}^{h} (\overline{x}_{j} - \overline{x}_{m})(\overline{y}_{j} - \overline{y}_{m})}{\sqrt{\sum_{j=0}^{h} (\overline{x}_{j} - \overline{x}_{m})^2} \sqrt{\sum_{j=0}^{h} (\overline{y}_{j} - \overline{y}_{m})^2}},
\end{equation}

where $\overline{x}_m=\frac{\sum_{j=0}^{h} \overline{x}_{j}}{h}$ is the arithmetic mean of hourly average traffic $\overline{x}_{j}$ of a group $x$ over $h$ hours. $\overline{x}_{j}$ is calculated by taking a mean of the traffic in hour $j\in h$ for $c$ number of cells in group $x$: $\overline{x}_{j}=\frac{\sum_{i=1}^{c} x_{i}^{j}}{c}$, where $x_{i}$ is a cell in group $x$. Similarly, $\overline{y}_{j}$ and $\overline{y}_{m}$ are calculated for group $y$. The value of $h$ depends on the availability of mobile traffic data. After calculating the correlation among all groups, they are categorized into \( K \) categories in a manner that maximizes the PCC. Groups within a category are combined to form a single cluster, resulting in a total of \( K \) clusters. As these clusters encompass cells with similar traffic patterns irrespective of their geographical regions. This approach not only reduces the computational requirements for model training but also ensures higher accuracy during live inference.

\begin{figure}[!t]
\begin{minipage}{0.5\textwidth}
\centering
\includegraphics[width=0.75\textwidth]{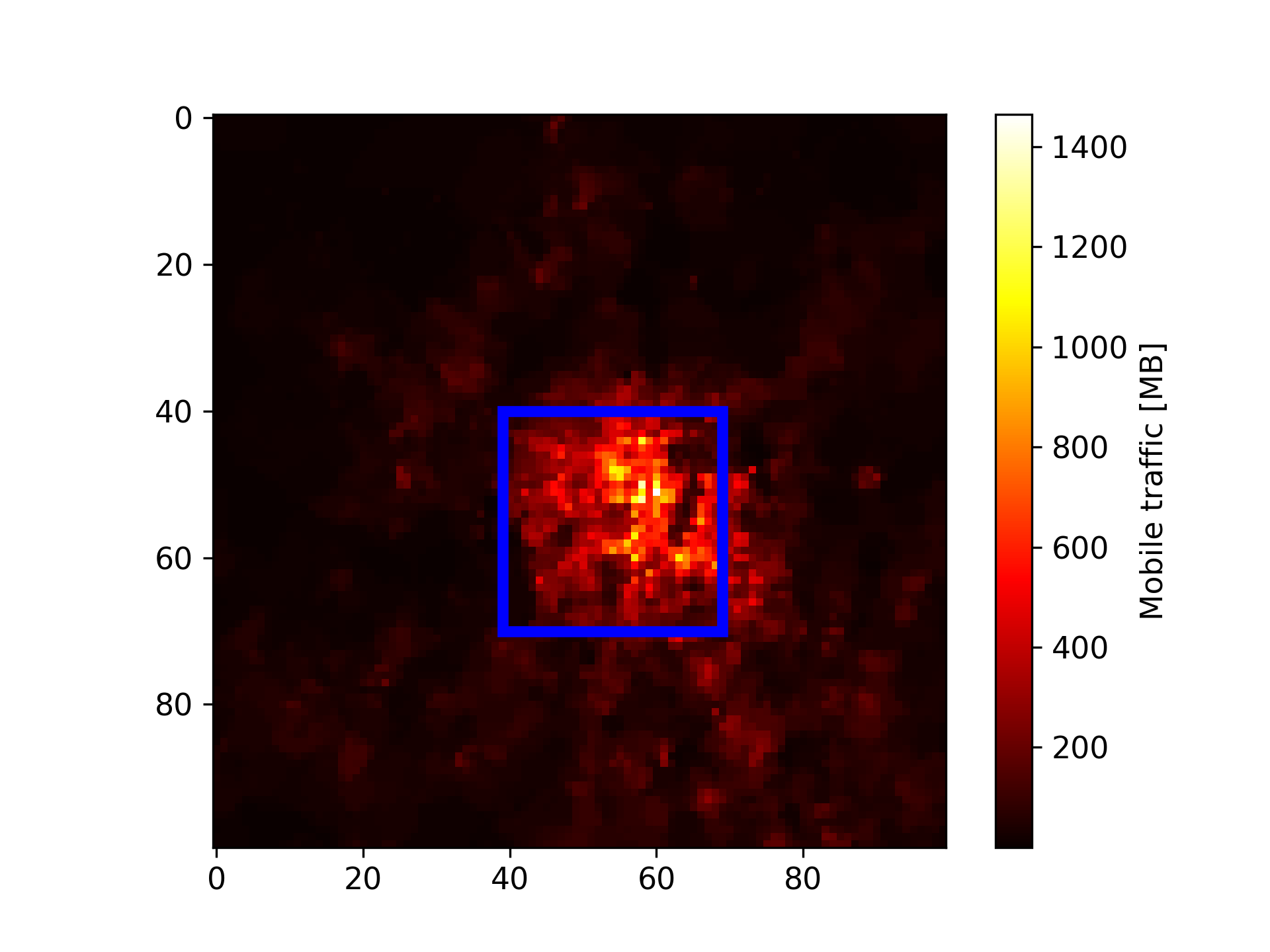}
\caption{100x100 average of mobile traffic heat map.}
\label{fig_3}
\end{minipage}
\end{figure}

\section{Performance Evaluation}

\subsection{Dataset and Evaluation Metrics}\label{AA}

The study evaluates an enhanced mobile traffic prediction using a dataset from Telecom Italia \cite{barlacchi2015multi}, which encompasses real-world mobile traffic data of Milan from November 1 to December 31, 2013. This data, spanning 10,000 cells, indicates in Fig.~\ref{fig_3} that the central 900 cells experience the highest traffic density. Given their significance in network operations, accurate traffic prediction for these cells is crucial. Accordingly, our focus is primarily on these 900 cells, in line with \cite{he2020graph}. For model training and cross-validation, data from November 1-30 is utilized. Specifically, the first 20 days are used for training the proposed multi TCN-LSTM model, while the last 10 days serve for cross-validation and evaluation.

The performance of the proposed method is evaluated using Mean Absolute Percentage Error (MAPE) and Mean Absolute Error (MAE). MAPE is calculated as follows:
\begin{equation}
MAPE = \frac{100}{n} \sum_{i=1}^{n} \left|\frac{q_i - \hat{q}_i}{q_i}\right|, %
\label{MAPE}
\end{equation}
where $n$ is the number of data points, $q_i$ is the actual value, and $\hat{q}_i$ is the predicted value. MAE is calculated as follows:
\begin{equation}
MAE = \frac{1}{n} \sum_{i=1}^{n} |q_i - \hat{q}_i|.
\end{equation}

\begin{figure}[!t]
\centerline{\includegraphics[width=0.4\textwidth]{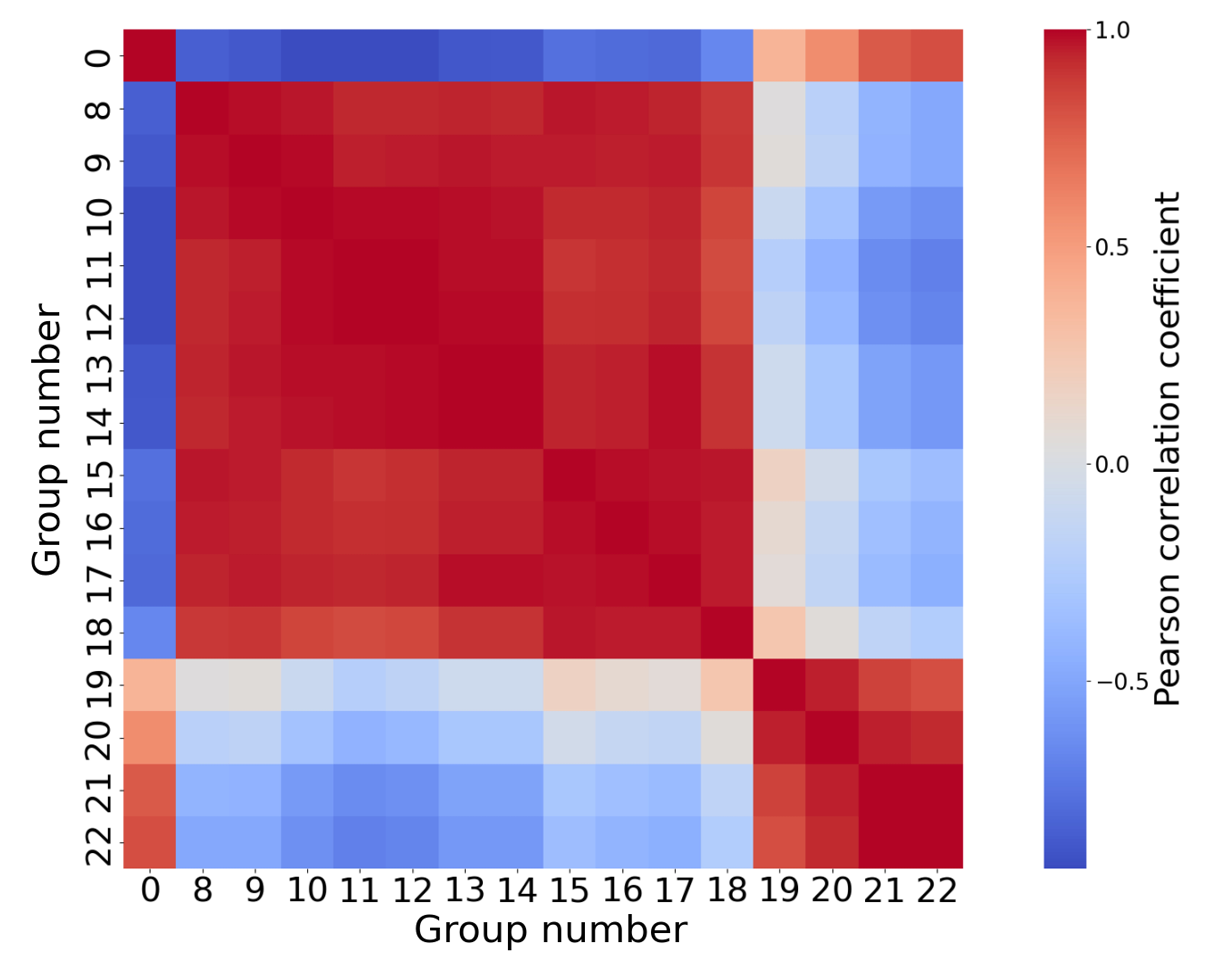}}
\caption{Pearson correlation coefficient heat map from groups.}
\label{fig_4}
\end{figure} 

\subsection{Experiment Results}

An individual multi TCN-LSTM model trains for each of the 24 initial groups defined. However, this is computationally infeasible for network operators. To reduce the number of groups by merging, further correlation among cells is determined through PCC which is depicted using heatmap in Fig.~\ref{fig_4}. Two distinct patterns emerge in the heatmap, where cells in groups 8-18 show a strong correlation presented with red boxes, and a strong correlation also exists between cells in groups 19-22 and 0. The blue color in the remaining heatmap exhibits that there is only a weak correlation between these two sets of clusters. Based on these observations, initial groups 8-18 are merged together to form a business hours cluster that represents business and commercial areas. Similarly, initial groups 19-22 and 0 are merged to form a leisure hours cluster that mainly represents residential areas. Through this process, 24 initial groups are reduced to two main clusters (i.e., $K$=2) which reduces the computation cost by 92\% and becomes computationally feasible for the operators.

\begin{table}[!t]
\caption{MAPE and MAE comparisons of the proposed model, state-of-the-art, and conventional DL methods.}
\begin{center}
\begin{tabular}{|c|c|c|}
\hline
\textbf{Model}&\textbf{MAPE}&\textbf{MAE} \\
\hline
LSTM & 0.2862 & 31.5985 \\
MLP & 0.1846 & 31.7576 \\
multi TCN-LSTM & 0.1608 & 30.3145 \\
LSTM-C & 0.1344 & 30.0124 \\
MLP-C & 0.1331 & 31.5198 \\
GASTN & 0.1760 & 29.8757 \\
\textbf{multi TCN-LSTM-C} & \textbf{0.1260} & \textbf{29.8568} \\
\hline
\end{tabular}
\label{tab2}
\end{center}
\end{table}

\begin{figure}[!t]
\centerline{\includegraphics[width=0.45\textwidth]{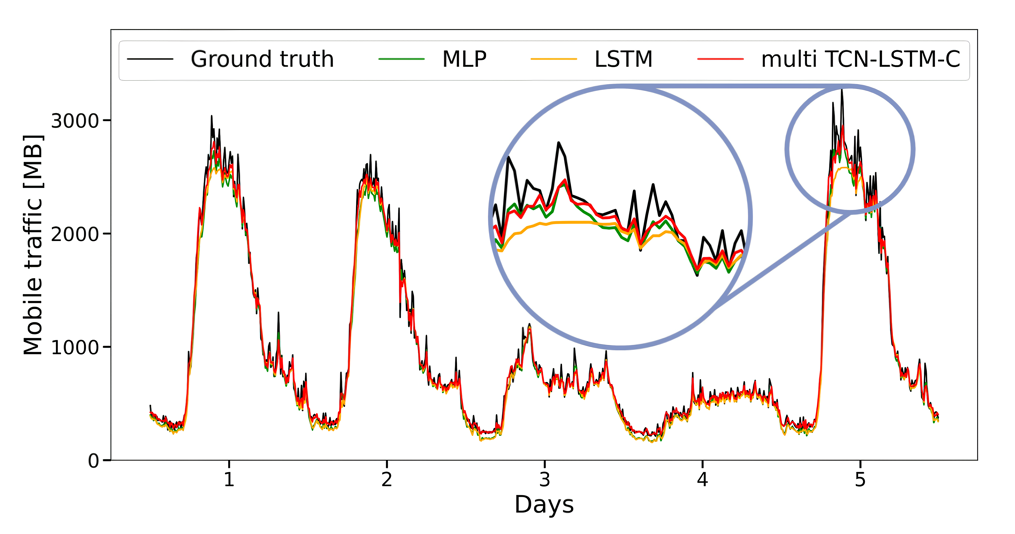}}
\caption{Ground truth and predicted mobile traffic values in cell 4956.}
\label{fig_5}
\end{figure}

The impact of the proposed clustering method is evaluated by comparing the performance of the presented multi TCN-LSTM and other conventional prediction models with and without clustering in Table \ref{tab2} where '-C' indicates clustering. The results show that MAPE and MAE of all models except Graph Attention spatioTemporal Network (GASTN) \cite{he2020graph} are reduced with clustering. In particular, MLP and LSTM methods exhibit a reduction in MAPE by 28\% and 53\%, respectively, along with 22\% reduction for multi TCN-LSTM when clustering is applied. This highlights the effectiveness of the proposed clustering method for not only the presented multi TCN-LSTM model but also for other methods while confirming its ability to comprehensively capture the geospatial and temporal correlations. Moreover, the substantially better performance of multi TCN-LSTM compared to LSTM and MLP even without clustering signifies its ability to intricately learn the underlying temporal features of mobile traffic. These distinctive characteristics of the proposed clustering method and multi TCN-LSTM enable them together to achieve 28\% reduced MAPE compared to GASTN and predict the mobile traffic more precisely. This is confirmed by the traffic prediction comparison depicted in Fig.~\ref{fig_5}, where it is evident that the proposed multi TCN-LSTM with clustering captures the features more adeptly compared to the other two models. Particularly during high mobile traffic periods, this distinction becomes more pronounced. This demonstrates the proposed model's capability to precisely predict the dynamic changes in mobile traffic over time and enable operators to devise more effective traffic engineering and resource management policies.
\section{Conclusion}
This paper contributes several insights for high-precision mobile traffic prediction in urban cities. It proposes a novel clustering mechanism along with multi TCN-LSTM model to fully exploit the geospatial and temporal features of the traffic. The performance of the proposed approach is evaluated against state-of-the-art and conventional DL methods using the open mobile traffic dataset of Milan. The MAPE and MAE results confirm the superiority of the proposed multi TCN-LSTM model over conventional MLP and LSTM models even without clustering. This signifies that the multi TCN-LSTM captures spatiotemporal characteristics more accurately than existing models. With clustering, the performance of multi TCN-LSTM as well as MLP and LSTM models improves by 22\%, 28\%, and 53\%, respectively in terms of MAPE, demonstrating the versatility of the proposed clustering technique and showcasing its applicability across various models. Moreover, the combination of the proposed clustering and the multi TCN-LSTM model surpassed the existing state-of-the-art studies by a 28\% increase in performance. For future work, we plan to refine the structure of RCCs in the multi TCN-LSTM to further enhance long-term memory and the capturing of spatiotemporal characteristics. Additionally, the proposed clustering technique will be applied to a more realistic scenario with 10,000 regions instead of 900, in order to evaluate its performance in larger contexts.

\section*{Acknowledgment}
This work was supported by the BK21 FOUR Project and by IITP grant funded by the Korea government(MSIT) under Artificial Intelligence Graduate School (No.2019-0-00421), and Artificial Intelligence Innovation Hub (No.2021-0-02068) and the ICT Creative Consilience Program (IITP-2023-2020-0-01821).

\bibliographystyle{IEEEtran} 
\bibliography{IEEEabrv, refer}

\begin{thebibliography}{1}
\providecommand{\url}[1]{#1}
\csname url@samestyle\endcsname
\providecommand{\newblock}{\relax}
\providecommand{\bibinfo}[2]{#2}
\providecommand{\BIBentrySTDinterwordspacing}{\spaceskip=0pt\relax}
\providecommand{\BIBentryALTinterwordstretchfactor}{4}
\providecommand{\BIBentryALTinterwordspacing}{\spaceskip=\fontdimen2\font plus
\BIBentryALTinterwordstretchfactor\fontdimen3\font minus \fontdimen4\font\relax}
\providecommand{\BIBforeignlanguage}[2]{{%
\expandafter\ifx\csname l@#1\endcsname\relax
\typeout{** WARNING: IEEEtran.bst: No hyphenation pattern has been}%
\typeout{** loaded for the language `#1'. Using the pattern for}%
\typeout{** the default language instead.}%
\else
\language=\csname l@#1\endcsname
\fi
#2}}
\providecommand{\BIBdecl}{\relax}
\BIBdecl

\bibitem{he2020graph}
K.~He, X.~Chen, Q.~Wu, S.~Yu, and Z.~Zhou, ``Graph attention spatial-temporal network with collaborative global-local learning for citywide mobile traffic prediction,'' \emph{IEEE Transactions on mobile computing}, vol.~21, no.~4, pp. 1244--1256, 2020.

\bibitem{rau2021forescating}
F.~Rau, I.~Soto, and D.~Zabala-Blanco, ``Forescating mobile network traffic based on deep learning networks,'' in \emph{2021 IEEE Latin-American Conference on Communications (LATINCOM)}.\hskip 1em plus 0.5em minus 0.4em\relax IEEE, 2021, pp. 1--6.

\bibitem{wan2022network}
X.~Wan, H.~Liu, H.~Xu, and X.~Zhang, ``Network traffic prediction based on lstm and transfer learning,'' \emph{IEEE Access}, vol.~10, pp. 86\,181--86\,190, 2022.

\bibitem{tiwari2022internet}
V.~Tiwari, C.~Pandey, and D.~S. Roy, ``Internet activity forecasting over 5g billing data using deep learning techniques,'' in \emph{2022 International Conference on Intelligent Controller and Computing for Smart Power (ICICCSP)}.\hskip 1em plus 0.5em minus 0.4em\relax IEEE, 2022, pp. 1--4.

\bibitem{sone2020wireless}
S.~P. Sone, J.~J. Lehtom{\"a}ki, and Z.~Khan, ``Wireless traffic usage forecasting using real enterprise network data: Analysis and methods,'' \emph{IEEE Open Journal of the Communications Society}, vol.~1, pp. 777--797, 2020.

\bibitem{zhang2021dual}
C.~Zhang, S.~Dang, B.~Shihada, and M.-S. Alouini, ``Dual attention-based federated learning for wireless traffic prediction,'' in \emph{IEEE INFOCOM 2021-IEEE conference on computer communications}.\hskip 1em plus 0.5em minus 0.4em\relax IEEE, 2021, pp. 1--10.

\bibitem{chen2023spatial}
X.~Chen, G.~Chuai, K.~Zhang, and W.~Gao, ``Spatial-temporal cellular traffic prediction: A novel method based on causality and graph attention network,'' in \emph{2023 IEEE Wireless Communications and Networking Conference (WCNC)}.\hskip 1em plus 0.5em minus 0.4em\relax IEEE, 2023, pp. 1--6.

\bibitem{barlacchi2015multi}
G.~Barlacchi, M.~De~Nadai, R.~Larcher, A.~Casella, C.~Chitic, G.~Torrisi, F.~Antonelli, A.~Vespignani, A.~Pentland, and B.~Lepri, ``A multi-source dataset of urban life in the city of milan and the province of trentino,'' \emph{Scientific data}, vol.~2, no.~1, pp. 1--15, 2015.

\bibitem{li2020prediction}
F.~Li, Z.~Zhang, Y.~Zhu, and J.~Zhang, ``Prediction of twitter traffic based on machine learning and data analytics,'' in \emph{IEEE INFOCOM 2020-IEEE Conference on Computer Communications Workshops (INFOCOM WKSHPS)}.\hskip 1em plus 0.5em minus 0.4em\relax IEEE, 2020, pp. 443--448.

\end{thebibliography}

\vspace{12pt}
\end{document}